\pdfoutput=1
\documentclass[11pt]{article}
\usepackage[margin=1in]{geometry}
\usepackage{graphicx}
\usepackage{booktabs}
\usepackage{tabularx}
\usepackage{amsmath,amssymb}
\usepackage[dvipsnames]{xcolor}
\usepackage[colorlinks=true,linkcolor=NavyBlue,citecolor=NavyBlue,urlcolor=NavyBlue]{hyperref}
\usepackage{caption}
\usepackage{array}
\newcolumntype{Y}{>{\raggedright\arraybackslash}X}
\newcommand{\Lpair}{\Delta L_{\text{pair}}}
\newcommand{\Lclone}{\Delta L_{\text{clone}}}
\newcommand{\jev}{\texttt{jev-1.13.0}}

\title{\textbf{Does a Hosted Decision Model Respect Irrelevant Alternatives?}\\[4pt]
\large A Pre-Registered Test of Option-Set Dependence in Jev}
\author{Dushyant Rajput \quad Nirdesh Chauhan \quad Siddharth Kosaraju\\[4pt]
{\normalsize AltSlate Labs LLP}\\[2pt]
{\small\texttt{dushyant@altslate.com} \quad \texttt{nirdesh@altslate.com} \quad \texttt{siddharth@altslate.com}}}
\date{October 2026}

\begin{document}
\maketitle

\begin{abstract}
Decision APIs that return a probability for each option are routinely used as if those probabilities were
properties of the options themselves. That reading is valid only if the model satisfies independence of
irrelevant alternatives (IIA): adding an option that is neither the right answer nor its strongest rival
should leave the odds between those two unchanged. We test this in a pre-registered study of
\jev{}, a hosted decision model, using 299,463 calls on intent classification (BANKING77, CLINC150)
and topic classification (AG News). \textbf{Every added option moves the log-odds} between the gold label and
the strongest wrong label: the root-mean-square systematic shift is 0.20--0.52 nats, 1.2 to 3 times the
run-to-run variation, even when the added option takes at most 0.1\% of the probability mass. The shifts are
directional. An option similar to the wrong answer draws probability from it (+0.10 nats toward the gold
label). A paraphrase of the wrong answer produces an intermediate pattern: the gold label's probability
is almost conserved, but the paraphrase receives only about a third of the wrong answer's share rather
than the half that strict substitution implies. The distortion grows steadily with the logarithm of the
number of options, reaching about 1 nat at 255 options, and off-domain padding pushes the odds against the
gold label. In contrast, questions packed into one call do not interact (bias $<0.001$). Scoring each
option with an independent yes/no question removes the option-set dependence but picks the right answer
8.7 points less often. Jev sits between two local baselines, an option-independent scorer and a prompt
readout. Its probabilities are best read as relative to the option list they were computed over.
\end{abstract}

\section{Introduction}

Hosted decision models expose classification as an API: the caller supplies a text, a question and a list of
options, and receives a probability for each option. In production these probabilities are thresholded,
logged, compared across calls and fed into downstream routing. Every one of those uses quietly assumes that
the probability an option receives depends on the text and the option, and not on which other options
happened to be listed. In choice theory this assumption is Luce's choice axiom, whose best-known consequence is
\emph{independence of irrelevant alternatives} (IIA)~\cite{luce1959}: the ratio $P(A)/P(B)$ does not depend on
the rest of the choice set.

IIA is a strong assumption even for people. Adding a near-duplicate of one option (Debreu's
counterexample~\cite{debreu1960}, known in transport demand as the ``red bus / blue bus'' problem), a similar
option~\cite{tversky1972}, or a dominated decoy~\cite{huber1982,simonson1989} all change human choice ratios.
For language models there is ample evidence of sensitivity to the \emph{order} and \emph{surface form} of
options~\cite{zhao2021,holtzman2021,pezeshkpour2024,zheng2024}. Whether a dedicated, calibrated decision model
keeps its pairwise odds fixed when the \emph{set} of options changes has, to our knowledge, not been measured
directly.

We ask that question of \jev{}~\cite{typesafe}, a hosted non-generative decision model that returns option
probabilities on a 0.01 grid. We pre-registered five hypotheses, froze all items, decoys and analysis code
before the first measurement call, and ran four experiments (284,463 calls plus a 15,000-call drift check)
at a total cost of \$6.79.

\paragraph{Contributions.}
\begin{itemize}
\item A pre-registered measurement design for IIA in probability-returning models. It separates systematic shift
  from run-to-run noise with an unbiased squared-excess statistic, controls position with full permutations,
  and discriminates Luce from elimination-by-aspects with a clone test (\S\ref{sec:design}).
\item Evidence that Jev violates IIA: every added option moves the gold-vs-rival log-odds beyond noise, with
  structured direction (similarity, partial substitution) and growth in $\log K$
  \mbox{(\S\ref{sec:exp1}--\ref{sec:exp2})}.
\item Evidence that the multi-question API is isolated: unrelated questions in the same call do not interact
  (\S\ref{sec:exp3}).
\item A tested mitigation (one yes/no question per option) that restores invariance at an accuracy cost
  (\S\ref{sec:exp4}), and two local baselines that place Jev between an option-independent scorer and a prompt
  readout (\S\ref{sec:baselines}).
\end{itemize}

\section{Background}

\paragraph{Luce, IIA and clones.} Under Luce's axiom each option $i$ has a positive strength $v_i$ and
\mbox{$P(i \mid S) = v_i / \sum_{j \in S} v_j$} for any set $S$; the multinomial logit model has the same
form~\cite{mcfadden1974}. The ratio $P(A)/P(B)$ is then fixed. The axiom makes a sharp, often wrong, prediction
for \emph{clones}: adding an exact copy $B'$ of $B$ halves the mass of the $\{B, B'\}$ ``family'' relative to
$A$, so $A$ loses probability to what is effectively the same option~\cite{debreu1960}. Elimination by aspects
(EBA)~\cite{tversky1972} predicts the opposite: the clone shares no distinguishing aspect with $A$, so $A$'s
probability is unchanged and $B$ and $B'$ split $B$'s former share. The clone therefore discriminates the two
models. In log-odds $L = \ln P(A)/P(B)$, Luce predicts $\Lpair = 0$ and a pooled shift
\mbox{$\Lclone = \ln P(A)/(P(B) + P(B')) - L = -\ln 2$}; EBA predicts $\Lpair = +\ln 2$ and $\Lclone = 0$.

\paragraph{Context effects.} Beyond clones, people show similarity effects (a new option takes share mainly from
options like it~\cite{tversky1972}), attraction to an option that asymmetrically dominates a
decoy~\cite{huber1982}, and compromise effects~\cite{simonson1989}; see~\cite{rieskamp2006} for a review. Our
decoy conditions (near-A, near-B) test the first two directions.

\paragraph{Language models as choosers.} LLM option probabilities are known to depend on option order and
labels~\cite{pezeshkpour2024,zheng2024}, on surface-form competition among strings with the same
meaning~\cite{holtzman2021}, and on prompt-level priors that calibration can partly remove~\cite{zhao2021}.
Multiple-choice prompting lists all the options in the prompt~\cite{robinson2023}, so each option's score can
depend on the others. We study a system offered specifically as a calibrated decision API, and we measure set
dependence directly rather than order dependence alone.

\section{Study design}\label{sec:design}

\begin{figure}[t]\centering
\includegraphics[width=0.95\textwidth]{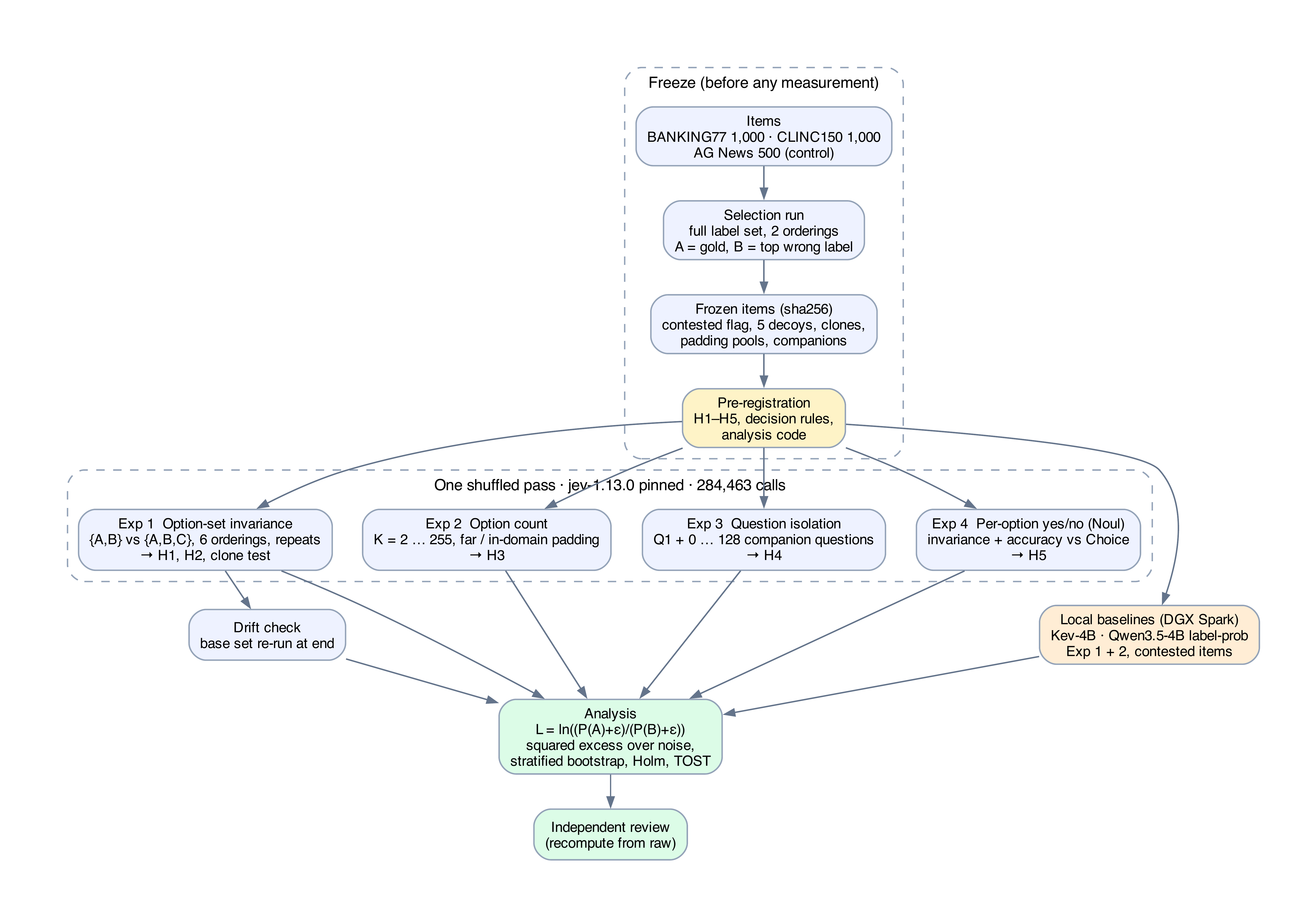}
\caption{Study workflow. Items, decoys, padding, companions and analysis code were frozen (sha256) and
pre-registered before any measurement call. All experiments ran in one shuffled pass with the model version
pinned; the base set was re-run at the end as a drift check. Local baselines ran Exp~1--2 on contested items (\S3.2).}
\label{fig:workflow}
\end{figure}

\subsection{System under test}

\jev{} is called through \texttt{POST /v1/systemone}~\cite{typesafe} with a text (\texttt{state}) and one or
more named questions. A \emph{Choice} question lists options (with optional descriptions) and returns a
probability per option. A \emph{Noul} question is a yes/no question returning $P(\text{yes})$. Probabilities lie
on a 0.01 grid and exact zeros occur (33\% of probabilities in the pilot). The model version is pinned on every
call, and any other returned version aborted the run; all 299,463 calls returned \jev{}. The API is not
deterministic: two identical calls typically differ (\S\ref{sec:exp1}).

\subsection{Items, the A/B pair and contested items}

We used BANKING77~\cite{casanueva2020} (77 banking intents; 1,000 test items, excluding the pilot rows),
CLINC150~\cite{larson2019} (150 intents, out-of-scope removed; 1,000 items) and AG News~\cite{zhang2015}
(4 topics; 500 items, a control). Option names are the label names in plain words. A \emph{selection run}
(the full label set, two random orderings, never used as a measurement) fixed, for each item, $A$ = the gold label and
$B$ = the wrong label with the highest mean probability. Selection-run accuracy was 0.82 / 0.91 / 0.87. Shift
analyses use \emph{contested} items, where $A$ and $B$ both received $\geq 0.05$: 319 BANKING77, 254 CLINC150 and
74 AG News (647). Accuracy and flip rates use all items.

\subsection{Measure and noise}

The outcome is the smoothed log-odds $L = \ln\bigl((P(A)+\varepsilon)/(P(B)+\varepsilon)\bigr)$ with $\varepsilon = 0.005$
(sensitivity: 0.01). Each $L$ is averaged over a matched pair of orderings (A before B, and B before A), and
both sides of every contrast use the same number of calls. For an added option $C$, the shift is
\mbox{$\Delta L_c = L(\{A,B,C\}) - L(\{A,B\})$}.

Because Jev is stochastic, $|\Delta L|$ is inflated by noise, and by different amounts for different option
sets. We therefore ran every option set twice and use the \emph{squared excess}
\begin{equation}
y = \Delta L_c^2 - \frac{n_c^2 + n_2^2}{2},
\end{equation}
where $n_c$ and $n_2$ are the repeat differences of the 3-option and 2-option sets. If each call is truth plus
independent noise, $\mathbb{E}[y]$ equals the squared systematic shift whatever the noise level of each set. We
report $\sqrt{\mathbb{E}[y]}$ as the \emph{RMS systematic shift}, and $\sqrt{(\mathbb{E}[n_c^2] + \mathbb{E}[n_2^2])/2}$, the noise level of the same contrast, as the \emph{RMS run-to-run noise}. On synthetic null data the statistic averaged
zero, whereas a naive $|\Delta L| - |\text{noise}|$ difference was biased upward.

\subsection{Experiments}

\begin{table}[t]\centering\small
\begin{tabularx}{\textwidth}{@{}l Y r@{}}
\toprule
\textbf{Experiment} & \textbf{Manipulation} & \textbf{Calls}\\
\midrule
Exp 1: one added option (2,500 items) & Base $\{A,B\}$ (both orders $\times$ 3 repeats) vs $\{A,B,C\}$ for five
  decoys $C$, each in all 6 orderings plus a repeat of the two $C$-last orderings & 115,000\\
Exp 1n: neutral names (200 items) & Options renamed \texttt{Option NN}, label text moved into the description & 4,800\\
Exp 2: option count (1,000 items) & $K \in \{2,4,\dots,128,255\}$ with off-domain or in-domain padding; $A,B$
  in positions 1--2 (primary) or randomly permuted & 96,000\\
Exp 3: question isolation (1,000 items) & Choice$\{A,B\}$ alone vs with 1--128 unrelated yes/no questions,
  a related question about $A$ or $B$, last position, or a leading question id; 3 repeats each & 30,000\\
Exp 4: yes/no per option (1,000 items) & One Noul question per option, all in one call, over the Exp 1 and
  in-domain $K$ option sets; plus a 300-item one-question-per-call reference & 38,663\\
\bottomrule
\end{tabularx}
\caption{The four pre-registered experiments (284,463 calls) and the neutral-name probe. All jobs were shuffled
together and run in one pass; the Exp~1 base set was then re-run as a drift check (15,000 calls).}
\label{tab:design}
\end{table}

\paragraph{Decoys.} For each item the five added options are: \emph{far}, a label from another domain (AG News
topics other than business for the intent sets; a BANKING77 label for AG News); \emph{random}, a random in-set
label; \emph{near-A} and \emph{near-B}, the in-set label most similar to $A$ (or $B$), by cosine similarity of
all-MiniLM-L6-v2~\cite{reimers2019,wang2020minilm} label embeddings; and \emph{B-clone}, a fixed paraphrase of
$B$'s name. Near-decoys that a reviewer judged a plausible correct answer were dropped (\S\ref{sec:limits}).
Clones whose embedding was nearer $A$ than $B$ (69 of 2,500) were excluded from the primary clone test.

\paragraph{Padding.} Off-domain padding draws from 281 DBpedia~\cite{lehmann2015} class names,
GoEmotions~\cite{demszky2020} labels and AG News topics, after removing any label with cosine $\geq 0.45$ to an
intent label. In-domain padding uses the set's other labels, then off-domain labels once those run out. Each
item has one fixed padding sequence, so its option sets are nested across $K$.

\subsection{Hypotheses and statistics}

\begin{table}[!htb]\centering\small
\begin{tabularx}{\textwidth}{@{}l Y Y@{}}
\toprule
\textbf{H} & \textbf{Claim} & \textbf{Decision rule}\\
\midrule
H1 & Adding one option moves $L$ beyond noise & Mean $y > 0$, Holm-adjusted~\cite{holm1979} $p < .05$, in
  any of 5 decoys $\times$ 2 main sets\\
H2 & The shift is directional & Signed mean $\Delta L_c$ 95\% CI excludes 0 (near-B $> 0$: similarity;
  $< 0$: attraction)\\
H2-clone & Luce vs EBA & Luce: $\Lpair$ 90\% CI within $\pm 0.10$ and $\Lclone$ 95\% CI covers $-\ln 2$; EBA: the mirror
  image; otherwise intermediate\\
H3 & The shift grows with $K$ & Per-item OLS slope of $y_K$ on $\log_2 K$; 95\% CI lower bound $> 0$\\
H4 & Questions in one call are isolated & Each condition: mean $\Delta P(A)$ 90\% CI within $\pm 0.02$ and
  per-item RMS shift $< 0.01$ (two one-sided tests, TOST~\cite{schuirmann1987})\\
H5 & Yes/no per option fixes it at no cost & (a) RMS shift $< 0.10$ for all decoys and (b) accuracy
  difference 95\% CI lower bound $> -0.02$\\
\bottomrule
\end{tabularx}
\caption{Pre-registered hypotheses and decision rules. BANKING77 and CLINC150 are the main sets; AG News is a
reported control outside the Holm family.}
\label{tab:hyp}
\end{table}

The unit is the item: repeats and orderings are averaged within item. Confidence intervals come from 10,000
bootstrap resamples of items~\cite{efron1979}, stratified by dataset, and equivalence uses the 90\% interval.
Accuracy gives $1/k$ credit when $k$ options tie at the top, since the 0.01 grid makes ties common.

A 200-item pilot (4,900 calls including its selection run, \$0.08) checked four go/no-go gates (contested share,
signal, version stability, throughput) and was excluded from all confirmatory tests. An independent reviewer,
given only the pre-registration, code and raw responses, recomputed the headline numbers before write-up.

\section{Results}

All five verdicts are unchanged at $\varepsilon = 0.01$. Table~\ref{tab:verdicts} summarises them.

\begin{table}[!htb]\centering\small
\begin{tabularx}{\textwidth}{@{}l l Y@{}}
\toprule
\textbf{H} & \textbf{Verdict} & \textbf{Key estimate (95\% CI)}\\
\midrule
H1 & Supported, 10/10 cells & RMS shift 0.20--0.52 nats; RMS run-to-run noise 0.15--0.17\\
H2 & Supported (near-B, clone, far, random) & near-B $+0.099$ $[0.073, 0.124]$; near-A $+0.008$ $[-0.019, 0.035]$\\
H2-clone & Intermediate & $\Lpair = +0.365$ $[0.335, 0.396]$; $\Lclone = -0.110$ $[-0.163, -0.060]$; $\lambda = 0.53^{a}$\\
H3 & Supported & squared-excess slope 0.134 $[0.113, 0.157]$ nats$^2$ per doubling of $K$\\
H4 & Supported, 9/9 conditions & all mean $\Delta P(A)$ within $\pm 0.0006$ (90\% CIs within $\pm 0.0013$);
  per-item RMS $< 0.003$\\
H5 & Not supported & (a) RMS $< 0.02$, passes; (b) accuracy $-0.087$ $[-0.105, -0.070]$, fails\\
\bottomrule
\end{tabularx}
\caption{Verdicts under the pre-registered rules. $^{a}\lambda = \Lpair/\ln 2$, the fraction of the EBA prediction.}
\label{tab:verdicts}
\end{table}

\subsection{One added option (Exp 1)}\label{sec:exp1}

\begin{figure}[!htb]\centering
\includegraphics[width=\textwidth]{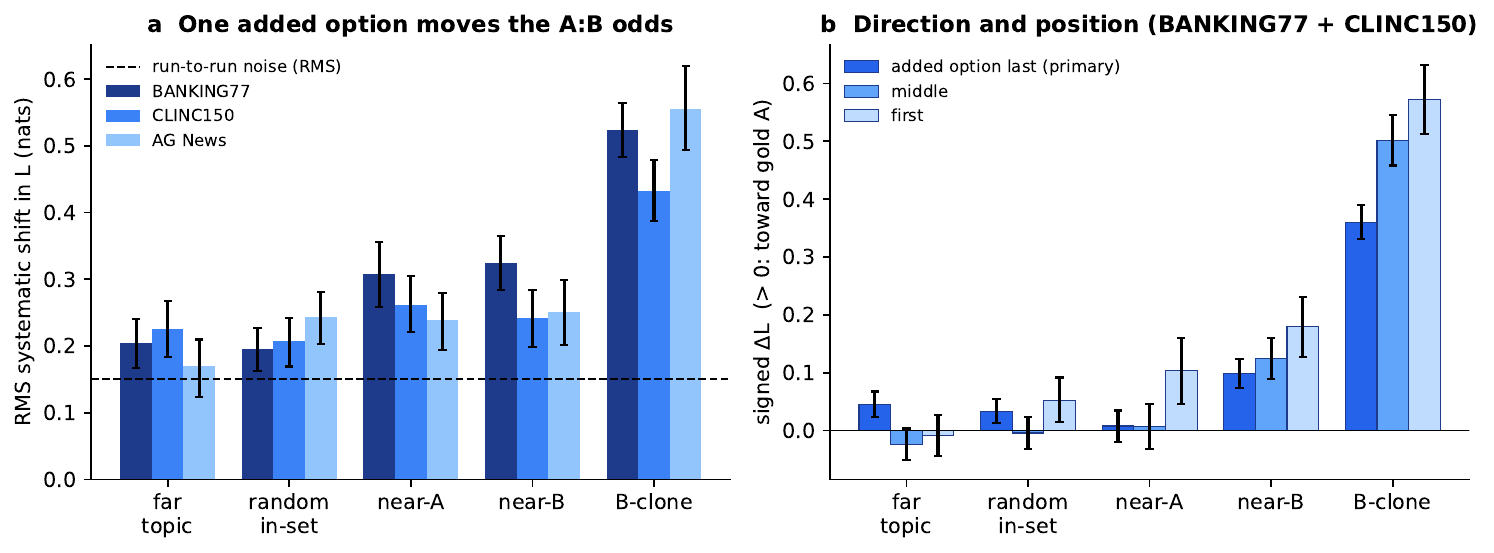}
\caption{One added option. (a) RMS systematic shift in the gold-vs-rival log-odds per decoy and dataset, with
95\% CIs; the dashed line is the RMS run-to-run noise between identical base calls. (b) Signed shift (positive =
toward the gold label) by the added option's position, BANKING77 and CLINC150 pooled.}
\label{fig:exp1}
\end{figure}

\paragraph{Every added option moves the odds.} As Fig.~\ref{fig:exp1}a shows, even an off-domain news topic, which takes at most 0.1\% of the probability, shifts $L$ by an RMS 0.20--0.23 nats in the main sets, against an RMS run-to-run noise
of 0.15--0.17 for the same contrasts (1.2--1.4 times). Similar options move it more (near-A 0.26--0.31, near-B
0.24--0.32; up to 2 times the noise) and the clone most (0.43--0.52; up to 3 times). All 10 main cells are
significant after Holm correction, and AG News behaves the same way (clone 0.56). Over all BANKING77 items, the
top choice in either added-option ordering differs from the base in 5--6\% of items for a far, random or near
decoy and 15\% for a clone; on contested items these rates are 14--16\% and 31\%.

\paragraph{Direction: similarity, not attraction.} With the added option listed last, the
conservative primary placement (Fig.~\ref{fig:exp1}b), a decoy similar to the wrong answer $B$ draws probability from $B$, moving the
odds toward the gold label by $+0.099$ $[0.073, 0.124]$. A decoy similar to the gold label does not detectably
hurt it ($+0.008$ $[-0.019, 0.035]$; this is a null not rejected, not an equivalence). The attraction
prediction, a near-B shift below zero, is rejected. Far and random decoys produce small positive shifts
($+0.046$, $+0.034$) that the reviewer found only where the base already favoured $A$, consistent with mild
sharpening rather than a decoy-specific mechanism; we treat their direction as exploratory.

\paragraph{Position adds to content.} For the similar decoys and the clone, the shift is larger when the added
option is listed first: near-B $+0.18$, near-A $+0.10$ and the clone $+0.57$, against $+0.10$, $+0.01$ and
$+0.36$ when listed last. For far and random decoys the signed shift is small at every position (between
$-0.02$ and $+0.05$). For similar options, the primary $C$-last contrast therefore understates what a caller with
arbitrary option order will see.

\paragraph{What does not drive it.} Over all items, Jev shows almost no first-option lean in mean probability
($P(A \mid A\text{ first}) - P(A \mid B\text{ first})$ between $-0.008$ and $-0.001$). On contested items the
lean in log-odds is small in the main sets ($-0.04$, $+0.06$) but larger in AG News ($-0.26$); the matched
orderings in every contrast cancel it. Replacing option names with neutral names (\texttt{Option 37}, with the
label text moved into the description) leaves the squared excess the same or larger (near-B 0.100 vs 0.085
named, on the same 200 items), so the effect is not a matter of the name strings alone. Accuracy and calibration
barely move with one added option (accuracy 0.855 base, 0.853--0.861 across decoys; expected calibration error 0.077 base).

\subsection{The clone: between Luce and elimination by aspects}\label{sec:clone}

\begin{figure}[t]\centering
\includegraphics[width=0.8\textwidth]{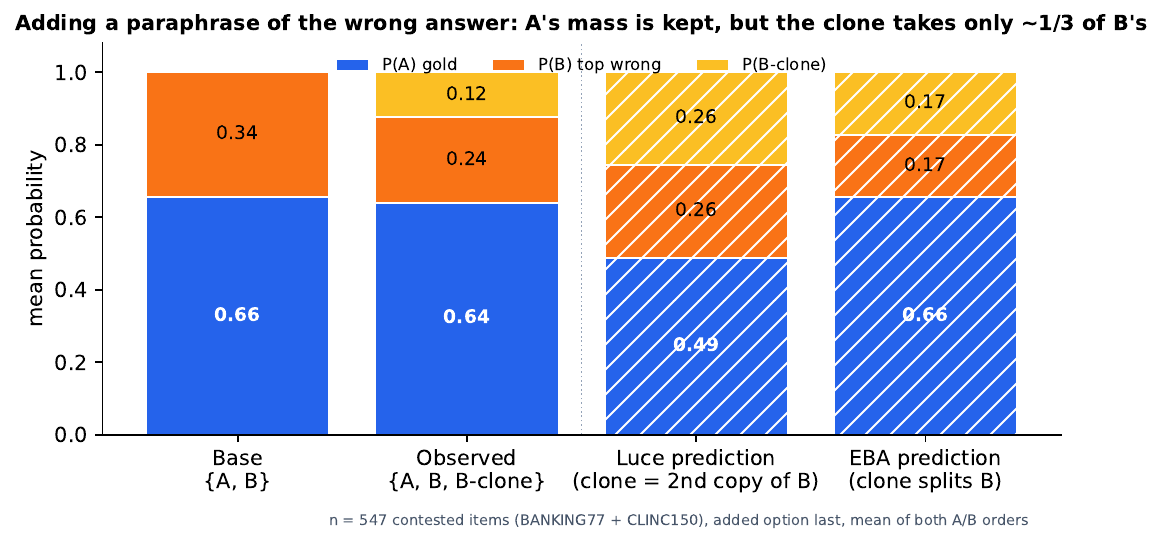}
\caption{Mean probabilities before and after adding a paraphrase of the wrong answer $B$ (547 primary contested
items, BANKING77 and CLINC150, clone listed last, both A/B orders), with the Luce and EBA predictions computed
from the base split.}
\label{fig:clone}
\end{figure}

The clone test rejects both pure models. In probability terms (Fig.~\ref{fig:clone}), the gold label keeps nearly
all its mass ($0.656 \to 0.640$, where Luce predicts $\approx 0.49$), as EBA predicts. But the paraphrase receives
only 0.124 of the B family's 0.360, about a third, where EBA predicts half. In log-odds the pairwise shift is
$\Lpair = +0.365$, a fraction $\lambda = 0.53$ of EBA's $\ln 2$, and the pooled shift is $\Lclone = -0.110$
rather than 0. On the pooled axis the result sits 84\% of the way from Luce to EBA. The two fractions differ because the pairwise and pooled axes are different nonlinear functions of the same mass split. Neither equivalence test passes, so the pre-registered verdict is \emph{intermediate}; including the 26 contested items whose clone was excluded gives
$\lambda = 0.52$. Jev treats a paraphrase as a partial substitute: it does not dilute the gold label, yet it does
not treat the paraphrase as equivalent to the original either. In 25\% of items the paraphrase even scores above
the original wording.

\subsection{Option count (Exp 2)}\label{sec:exp2}

\begin{figure}[!htb]\centering
\includegraphics[width=\textwidth]{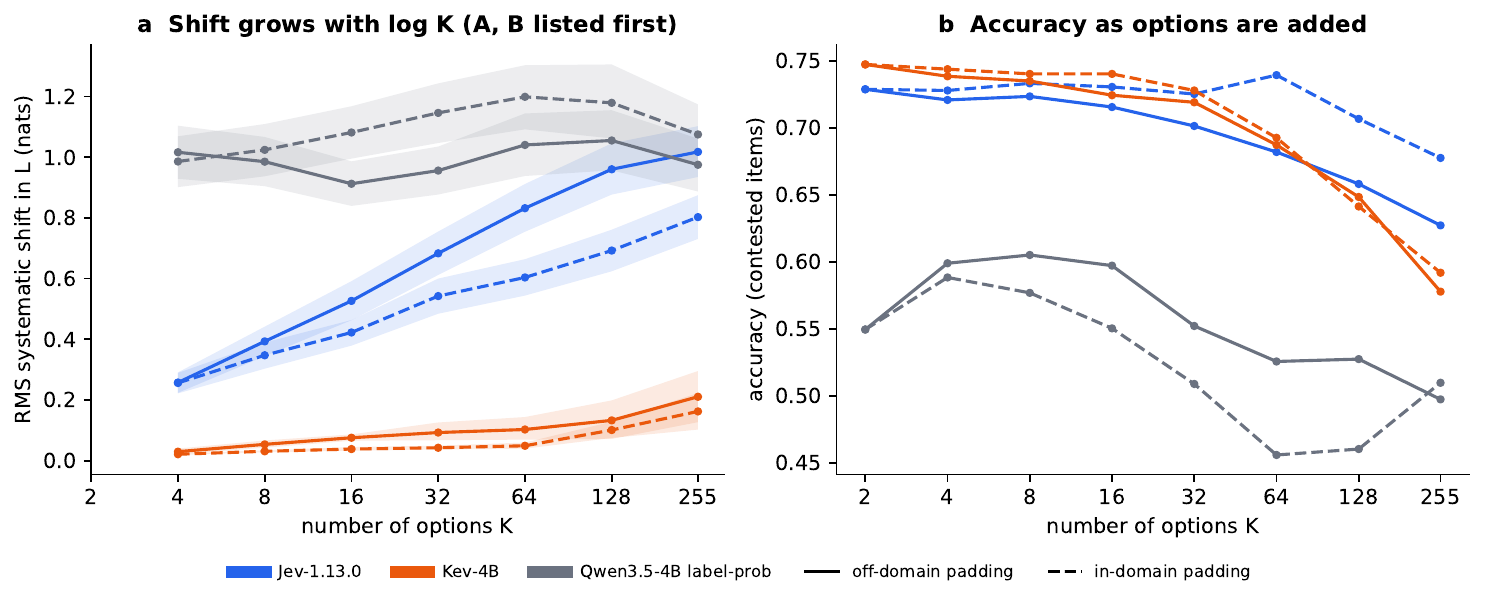}
\caption{Option count, 283 contested items, $A$ and $B$ listed first. (a) RMS systematic shift against $K$ (95\%
CI bands) for Jev and the two baselines. (b) Accuracy on the same items.}
\label{fig:exp2}
\end{figure}

The shift grows steadily with $\log_2 K$ with no plateau (Fig.~\ref{fig:exp2}). With off-domain padding the RMS
shift is 0.26 at $K = 4$, 0.68 at 32 and 1.02 at 255; with in-domain padding, 0.26, 0.54 and 0.80. The per-item
slope of the squared excess on $\log_2 K$ is 0.134 $[0.113, 0.157]$, steeper for off-domain padding (0.17) than
in-domain (0.09; difference 0.08), while run-to-run noise itself grows only slightly (the mean repeat difference rises by 0.006 nats per doubling).
Off-domain padding also turns the shift \emph{against} the gold label (signed $-0.31$ at $K = 255$), takes 6.8\%
of the mass and lowers accuracy from 0.853 to 0.814 over all items; in-domain padding costs little accuracy up to
$K = 64$ (0.859) and 1.5 points at 255. An exploratory segmented fit places a bend at $K = 16$. The secondary
mixed model did not converge, so we rely on the per-item slopes.

\subsection{Question isolation (Exp 3)}\label{sec:exp3}

\begin{figure}[!htb]\centering
\includegraphics[width=0.72\textwidth]{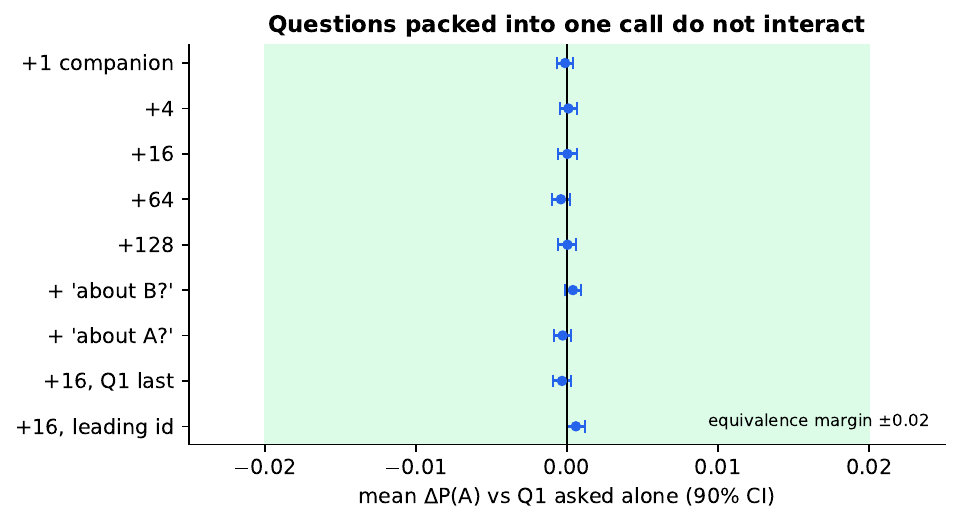}
\caption{Mean change in $P(A)$ when other questions share the call, with 90\% CIs, against the pre-registered
$\pm 0.02$ equivalence margin.}
\label{fig:exp3}
\end{figure}

In sharp contrast with the option set, the question set does not matter (Fig.~\ref{fig:exp3}). Asking
Choice$\{A,B\}$ together with 1 to 128 unrelated yes/no questions, with a related question about $A$ or about
$B$, in last position, or under a leading question id (\texttt{answer\_is\_<A>}) changes mean $P(A)$ by less than 0.0006 in every condition (90\% CIs within $\pm 0.0013$), and the per-item RMS shift is below 0.003. All nine conditions pass both equivalence
criteria. The only variation is noise: in 39.5\% of items the three identical repeats are not all equal.

\subsection{Yes/no per option (Exp 4)}\label{sec:exp4}

\begin{figure}[!htb]\centering
\includegraphics[width=\textwidth]{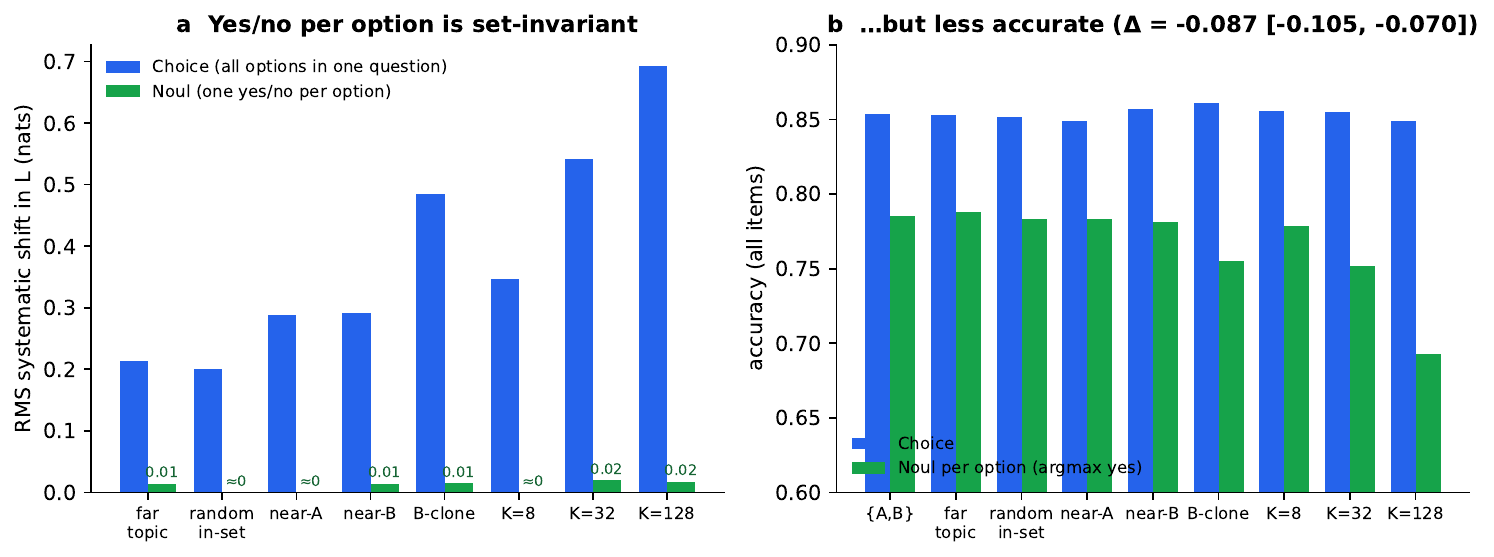}
\caption{One yes/no question per option. (a) RMS shift in the ratio of yes-probabilities under added options and
padding, against the same quantity for Choice. (b) Accuracy of Choice and of the argmax yes-probability on the
same option sets (y-axis starts at 0.6).}
\label{fig:exp4}
\end{figure}

Asking one yes/no question per option, all in one call, makes the scores invariant (Fig.~\ref{fig:exp4}a): no
decoy or padding up to $K = 128$ moves the yes-probability ratio by more than RMS 0.02, and per-call noise is
small (median $|\Delta L|$ 0.011). Asking the options together or in separate calls differs by less than the
separate calls' own repeat noise (0.043 vs 0.047). H5(a) passes. But taking the option with the highest
yes-probability as the answer is less accurate than Choice on identical option sets (Fig.~\ref{fig:exp4}b):
$-0.087$ $[-0.105, -0.070]$ overall, widening to 0.693 vs 0.849 at $K = 128$. H5(b) fails. The gap is not a tie
artefact: yes-probabilities tie at the top in only 3.9\% of calls, and crediting every tie in full still gives
0.785 against 0.854 for Choice.

\subsection{Baselines}\label{sec:baselines}

\begin{figure}[!htb]\centering
\includegraphics[width=0.72\textwidth]{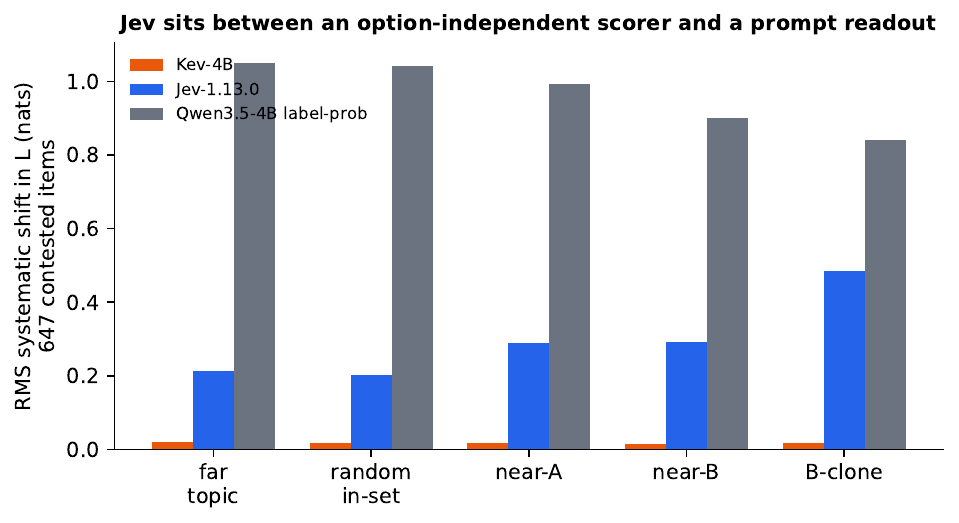}
\caption{RMS shift from one added option, by system (BANKING77 and CLINC150 contested items).}
\label{fig:baselines}
\end{figure}

Two local baselines ran Exp~1 on the 647 contested items and Exp~2 on the 283 contested Exp~2 items.
Kev-4B~\cite{kev4b} scores each option independently and then normalises. A Qwen3.5-4B-Base~\cite{qwen35}
label-probability readout lists all options in the prompt and reads the sequence probability of each option
name. Both are deterministic.

\begin{table}[!htb]\centering\small
\begin{tabular}{@{}l r r r@{}}
\toprule
 & \textbf{Jev-1.13.0} & \textbf{Kev-4B} & \textbf{Label-prob}\\
\midrule
Run-to-run noise (median $|\Delta L|$) & 0.09--0.11 & 0 & 0\\
RMS shift, one added option & 0.20--0.52 (AG 0.56) & 0.01--0.03 & 0.70--1.10\\
Signed shift, near-B / clone & $+0.10$ / $+0.36$ & $+0.004$ / $+0.005$ & $+0.18$ / $+0.02$\\
Clone $\Lpair$ / $\Lclone$$^{a}$ & $+0.37$ / $-0.11$ & $+0.005$ / $-0.56$ & $-0.01$ / $-0.43$\\
Contested accuracy, $K=2$ & 0.729 & 0.747 & 0.549\\
\quad $K=255$ off-domain / in-domain & 0.627 / 0.678 & 0.578 / 0.592 & 0.497 / 0.510\\
\bottomrule
\end{tabular}
\caption{Jev against the two baselines on the same contested items. Accuracy here is on contested items only
and is not comparable with the all-item figures above. $^{a}$On the 547 primary clone items.}
\label{tab:baselines}
\end{table}

Kev-4B is close to Luce, as an option-independent scorer must be: its pairwise odds barely move (RMS
0.01--0.03), and a clone dilutes the gold label ($\Lclone = -0.56$, near $-\ln 2 = -0.69$; strictly, the CI
$[-0.59, -0.53]$ excludes it). The label-probability readout is far less stable than Jev: one added option moves
its odds by about 1 nat, because the option list is part of the prompt. Jev sits between the two
(Fig.~\ref{fig:baselines}). On accuracy the ordering differs: Kev-4B degrades fastest as options are added
($0.747 \to 0.578$ off-domain), because an option-independent scorer gives every distractor its own share, while
Jev holds up best ($0.729 \to 0.627$ off-domain; 0.678 in-domain).

\subsection{Robustness}

All verdicts hold at $\varepsilon = 0.01$ (e.g.\ near-B $+0.093$, clone $\lambda = 0.51$, H3 slope 0.123
$[0.103, 0.143]$). Restricting to items with $P(B) \geq 0.10$ in both base calls (the pre-registered sensitivity
filter; 677 items) gives the same pattern (RMS shifts 0.21--0.30 for far, random and near decoys; 0.61 for the
clone). The end-of-run drift check matches the repeat noise (mean $\Delta L = -0.001$, mean $|\Delta L|$ 0.115 vs
0.119), so the pause in the run (\S\ref{sec:limits}) introduced no drift.

\section{Discussion}\label{sec:discussion}

\paragraph{What the pattern says about the mechanism.} Three findings constrain how Jev forms its probabilities.
First, options that take almost no mass still move the pairwise odds, so the coupling is not only normalisation
of independent scores, which would leave the ratio fixed (as Kev-4B shows). Second, the shift grows with
$\log K$ and is larger for unrelated padding. Third, the effect survives neutral names. These are consistent with
a model that encodes the option list jointly, with each option's score conditioned on the others. The
question-isolation result shows that this coupling is confined to a question's own options and does not leak
across questions. We did not test the mechanism directly; these are inferences from behaviour
(\S\ref{sec:future}).

\paragraph{How large is it in practice?} The shifts are modest relative to the decisions most callers make: one
added option rarely flips the top choice over all items (5--6\% for a non-clone decoy), and accuracy is stable up
to a few dozen in-domain options. They are large relative to what calibrated probabilities are used for: an RMS
shift of 0.3 nats moves a 0.70 probability to roughly 0.63--0.76, and the 1-nat shift at 255 options to roughly
0.46--0.86. Any use that compares probabilities across calls with different option lists, such as thresholds
tuned on one label set and applied to another, inherits this error.

\noindent\begin{minipage}{\textwidth}
\paragraph{Guidance for practitioners.}
\begin{enumerate}
\item Compare probabilities only within one option set; re-tune thresholds when the option list changes.
\item Remove near-duplicate options: a paraphrase splits its original's mass without diluting the rival.
\item Never pad with off-domain options: they push the odds against the gold answer and cost accuracy.
\item Use Choice to pick an answer and per-option yes/no questions to obtain comparable scores.
\item Pack independent questions into one call freely; there is no measurable interaction.
\item Average over repeats and both orderings when the probabilities matter: identical calls differ by about 0.1
  nats and about 1\% of top choices flip.
\end{enumerate}
\end{minipage}

\section{Limitations and deviations}\label{sec:limits}

\begin{itemize}
\item \textbf{One model version, one task family.} Results apply to \jev{} on single-label text classification
  with label names only. Later versions, other tasks and options with rich descriptions may differ.
\item \textbf{LLM-reviewed materials.} No human reviewer was available. The near-decoy check was done blind by an
  LLM (Claude) on 300 near-decoys, which flagged 4 (1.3\%); by extrapolation about 15 unreviewed
  plausible-correct near-decoys may remain among contested items, which would slightly inflate the near-decoy
  shifts. The clone paraphrases were also LLM-drafted.
\item \textbf{Contested items.} Shift analyses use items where both $A$ and $B$ receive at least 0.05, about a
  quarter of all items. Effects on uncontested items are not estimated.
\item \textbf{The run was paused once.} The account ran out of credits (HTTP 402) after 210,505 of 284,463 calls
  and resumed about six hours later. Jobs were shuffled across all experiments, so the pause is not confounded
  with condition, and the drift check is null.
\item \textbf{Baselines are exploratory.} They ran on contested items only, in bf16, and the label-probability
  readout used one job per forward pass because batching shifted scores by up to 0.06.
\item \textbf{One post-run analysis change.} A descriptive figure (3-option noise median) was first computed over
  all items instead of contested ones and was corrected. No tested quantity changed.
\item \textbf{Data copies.} BANKING77 was loaded from the \texttt{mteb/banking77} copy because the original
  loader no longer works.
\end{itemize}

\section{Future work}\label{sec:future}

\paragraph{Mechanism.} Refit each call's output as independent per-option scores normalised over the set; how
much of the shift this removes measures the share due to normalisation alone versus joint encoding.

\paragraph{Restoring invariance without losing accuracy.} Calibrate per-option yes/no scores against Choice on
held-out data, use them to break Choice near-ties, or average Choice over random option subsets to estimate
set-free scores. Each candidate should be judged on both criteria used here: shift (H1/H3) and accuracy (H5b).

\paragraph{The clone.} Vary paraphrase closeness from exact duplicate to related concept, and fit intermediate
substitution models (e.g.\ nested logit~\cite{mcfadden1978}) to estimate how substitutable Jev treats two
options.

\paragraph{Wider scope.} Multi-label and preference tasks, longer inputs, options with descriptions, non-English
text, asymmetric dominance and compromise designs~\cite{huber1982,simonson1989}, a regularity test (whether
adding an option can ever raise another's probability), and a human comparison on the same option sets. The
frozen pipeline makes re-running on later Jev versions cheap (about \$7).

\section{Conclusion}

\jev{} does not respect irrelevant alternatives. Every option added to a Choice question moves the odds between
the gold label and its strongest rival beyond run-to-run noise. The movement is structured: similar options draw
from the rival, paraphrases split their original's mass without diluting the gold label, and the distortion
grows with every doubling of the list. Questions sharing a call, in contrast, are isolated, and scoring each
option with its own yes/no question restores invariance at a cost of about nine accuracy points. Jev's
probabilities are therefore meaningful \emph{relative to the option set they were computed over} and should not
be read as fixed properties of the options. Practitioners can keep the benefits of the API by fixing the option
list, removing near-duplicates and choosing Choice or per-option yes/no according to whether they need a
decision or a comparable score.

\section*{Acknowledgments}

Study design, analysis code and drafting were assisted by Claude (Anthropic); an independent Claude reviewer
recomputed the headline numbers from the raw responses. Errors are the authors'.

\bibliographystyle{unsrt}
\bibliography{refs}

\clearpage
\appendix

\section{Exp 1 per-condition detail}

\begin{table}[h]\centering\small
\begin{tabular}{@{}l c r r r r@{}}
\toprule
\textbf{Added option} & \textbf{RMS shift} & \textbf{Signed $\Delta L$} & \textbf{C first} & \textbf{Flips} & \textbf{Mass to C}\\
 & \textbf{b77 / clinc / ag} & \textbf{(95\% CI)} & & \textbf{(b77)} & \textbf{(b77)}\\
\midrule
far topic & 0.20 / 0.23 / 0.17 & $+0.046$ $[0.024, 0.068]$ & $-0.009$ & 5.6\% & 0.1\%\\
random in-set & 0.20 / 0.21 / 0.24 & $+0.034$ $[0.013, 0.055]$ & $+0.053$ & 5.8\% & 0.1\%\\
near-A & 0.31 / 0.26 / 0.24 & $+0.008$ $[-0.019, 0.035]$ & $+0.104$ & 5.3\% & 1.0\%\\
near-B & 0.32 / 0.24 / 0.25 & $+0.099$ $[0.073, 0.124]$ & $+0.179$ & 5.4\% & 1.3\%\\
B-clone & 0.52 / 0.43 / 0.56 & $+0.360$ $[0.330, 0.389]$ & $+0.572$ & 14.7\% & 10.4\%\\
base repeat (noise) & 0.11 / 0.09 / 0.10$^{a}$ & --- & --- & 1.0\% & ---\\
\bottomrule
\end{tabular}
\caption{Exp~1 by condition. Signed shifts pool BANKING77 and CLINC150 (added option last). ``C first'' is the
signed shift with the added option listed first. Flips: top choice differs from the base. $^{a}$Median
$|\Delta L|$ between identical calls.}
\label{tab:exp1}
\end{table}

\section{Exp 2 curve}

\begin{table}[h]\centering\small
\begin{tabular}{@{}l r r r r r r r@{}}
\toprule
$\mathbf{K}$ & 4 & 8 & 16 & 32 & 64 & 128 & 255\\
\midrule
RMS shift, off-domain & 0.26 & 0.39 & 0.53 & 0.68 & 0.83 & 0.96 & 1.02\\
RMS shift, in-domain & 0.26 & 0.35 & 0.42 & 0.54 & 0.60 & 0.69 & 0.80\\
Signed $\Delta L$, off-domain & $+0.06$ & $+0.01$ & $-0.04$ & $-0.15$ & $-0.24$ & $-0.32$ & $-0.31$\\
Accuracy, off-domain & .852 & .853 & .852 & .847 & .840 & .830 & .814\\
Accuracy, in-domain & .855 & .856 & .856 & .855 & .859 & .849 & .839\\
Mass to padding, off-domain & 0.2\% & 0.2\% & 0.4\% & 0.8\% & 1.8\% & 3.8\% & 6.8\%\\
\bottomrule
\end{tabular}
\caption{Exp~2, $A$ and $B$ listed first. Shifts on 283 contested items; accuracy on all 1,000 items ($K = 2$:
0.853 off-domain, 0.854 in-domain).}
\label{tab:exp2}
\end{table}

\section{Reproducibility}

The pre-registration, frozen items (sha256 \texttt{cd5db21b}\ldots), analysis code and raw responses (one record
per call) are available from the authors on request. The main run
used 161.7M input tokens at \$0.042 per million (\$6.79) and ran at about 47 requests per second at concurrency
16. Baselines ran on one NVIDIA GB10 (DGX Spark): Kev-4B through its TypeSafe-compatible server, and the
label-probability readout with a shared-prefix branching forward pass, one job per pass.

\end{document}